\documentclass{article}

\usepackage{hyphenat}
\usepackage[main,final]{neurips_2026}

\usepackage[utf8]{inputenc}
\usepackage[T1]{fontenc}
\usepackage{url}
\usepackage{graphicx}
\usepackage{booktabs}
\usepackage{float}
\usepackage{amsmath}
\usepackage{amsfonts}
\usepackage{amssymb}
\usepackage{nicefrac}
\usepackage{microtype}
\usepackage[dvipsnames]{xcolor}
\usepackage{pifont}
\usepackage{capt-of}
\usepackage{hyperref}
\usepackage[capitalize,noabbrev]{cleveref}

\newcommand{\eg}{\emph{e.g.}}

\title{Motion Forcing: Decoupling Ego and Object Motion via Sparse Inputs for Structured Video Generation}

\author{%
  Tianshuo Xu\textsuperscript{1} \enspace
  Zhifei Chen\textsuperscript{1} \enspace
  Leyi Wu\textsuperscript{1} \enspace
  Hao Lu\textsuperscript{1} \enspace
  Ying-cong Chen\textsuperscript{1,2}\thanks{Corresponding author.} \\
  \textsuperscript{1}The Hong Kong University of Science and Technology (Guangzhou) \\
  \textsuperscript{2}The Hong Kong University of Science and Technology \\
  \texttt{\{txu647,zchen379\}@connect.hkust-gz.edu.cn} \\
  \\
  Project Page: \url{https://tianshuo-xu.github.io/Motion-Forcing/}%
}

\begin{document}

\maketitle

\vspace{-2em}
\begin{center}
    \centering
    \includegraphics[width=\linewidth]{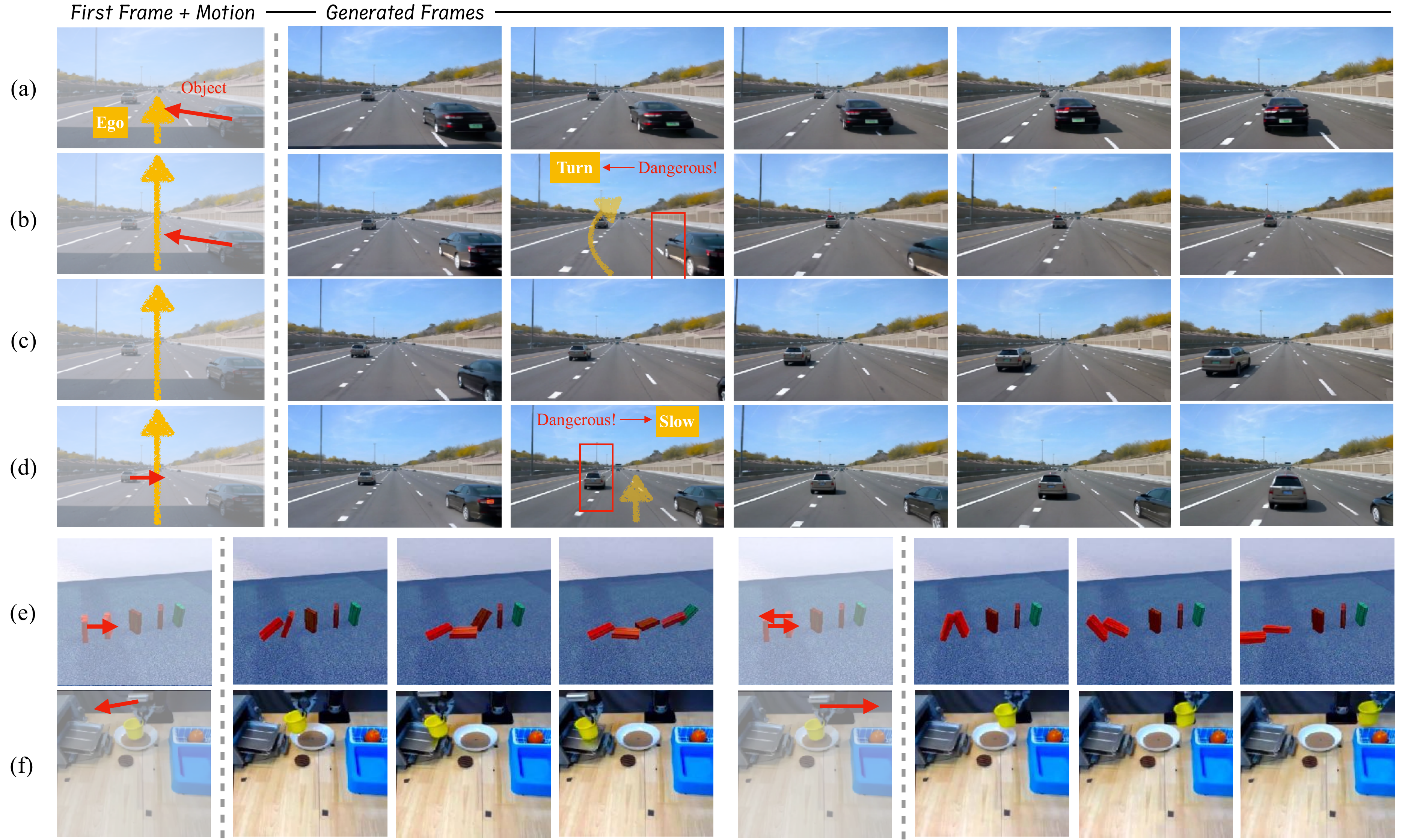}
    \captionof{figure}{\textbf{Motion Forcing} takes user input motion and the first frame to generate motion-coherent future frames. (a-d) show reactive ego-trajectories responding to dangerous scenarios initiated by other vehicles. (e) shows a physics scene where different actions lead to different collision results. (f) shows embodied AI, where directional inputs control a robotic hand to move an object in corresponding directions.}
    \vspace{1em}
    \label{fig:teaser}
\end{center}

\begin{abstract}
Structured video generation (e.g., autonomous driving) faces a fundamental domain gap between simple human control and dense physical reality. Text-to-Video models accept accessible inputs but struggle with precise spatial guidance, while dedicated autonomous driving world models require expensive, dense 3D annotations that are difficult for users to provide. To bridge this gap, we introduce \textbf{Motion Forcing}, a structured video generation framework that achieves precise multi-agent control using extremely sparse 2D inputs. Our key insight is to explicitly bridge human intent and visual synthesis via a hierarchical \textbf{``Point-Shape-Appearance''} paradigm. This approach decomposes generation into verifiable stages: modeling user intent as sparse geometric points, expanding them into dynamic depth maps to explicitly resolve 3D geometry and decouple ego-motion from object dynamics, and finally rendering high-fidelity textures. Furthermore, to elevate the model from passive instruction-following to active physical reasoning, we employ a \textbf{Masked Point Recovery} strategy. By forcing the reconstruction of dynamic depth from occluded trajectories, the model learns to infer causal continuations from partial observations. Extensive experiments demonstrate that Motion Forcing significantly outperforms state-of-the-art baselines on large-scale autonomous driving benchmarks. Additional validation in rigid-body physics and robotic manipulation confirms the structural integrity and generalizability of our framework.
\end{abstract}

\section{Introduction}
\label{sec:intro}

The development of realistic ``world models'' has become a critical foundation for embodied AI and autonomous driving~\cite{wang2023drivedreamer, zhao2024drivedreamer2, li2023drivingdiffusion, hu2023gaia, yang2024genad}. Unlike open-domain video generation designed for entertainment~\cite{zhuang2024vlogger, MakePixelsDance, zhang2024physdreamer}, world models operate in safety-centric domains where generating visually plausible textures is insufficient. These models must strictly adhere to fundamental physical laws---such as inertia, multi-agent collision dynamics, and object permanence---to support reliable decision-making and simulation~\cite{kang2024farvideogenerationworld, bansal2024videophy, meng2024physicalbench}. However, despite achieving remarkable perceptual fidelity, recent foundation models~\cite{chen2024videocrafter2, hong2022cogvideo, svd, yang2024cogvideox, yin2023dragnuwa} frequently suffer from physical hallucinations when generating complex interactive traffic scenarios, highlighting a profound gap between visual realism and rigorous 3D physical consistency.

We attribute this structural fragility to a massive generative domain gap between simple human intent and dense physical reality. Current methodologies cluster at two extremes, failing to bridge this void. On one end, text-to-video foundation models~\cite{sora2024, yang2024cogvideox, wan2025wan21} provide accessible interfaces (close to human intent) but lack the fine-grained spatial guidance required to orchestrate precise multi-agent interactions. On the other end, dedicated autonomous driving world models (e.g., MagicDrive3D~\cite{gao2024magicdrive3d}, DriveDreamer~\cite{wang2023drivedreamer}) achieve precise control but rely on dense, expensive 3D bounding boxes and HD maps. While these dense conditions closely approximate the final 3D output, they represent a prohibitive barrier for simple human interaction. Furthermore, attempting to bridge this gap using 2D motion intermediates, such as predicting dense optical flow from sparse trajectories (e.g., MoFA-Video~\cite{niu2024mofa}), fundamentally fails to capture 3D spatial structures. This lack of 3D awareness introduces severe entanglement between ego-motion and object dynamics, culminating in geometric distortions and physical collapse during complex interactions.

To explicitly bridge the semantic gap between accessible human input and complex physical output, we propose \textbf{Motion Forcing}, a decoupled structured video generation framework tailored for autonomous driving. Inspired by Diffusion Forcing~\cite{chen2024diffusion}, we bridge this generative void via a hierarchical \textbf{``Point-Shape-Appearance''} paradigm. To ensure intuitive control, we abstract human intent into extremely sparse \textbf{Points} carrying positional and scale attributes. Rather than forcing the network to hallucinate pixels directly from these 2D signals, we introduce an intermediate \textbf{Shape} stage, expanding points into dynamic depth maps. This intermediate representation serves as the perfect bridge: it explicitly encodes 3D structural geometry, remains entirely invariant to high-frequency RGB details, and provides a dense, pixel-aligned blueprint for the final output. This strictly decouples ego and object motion in a unified metric space. Finally, we render high-fidelity RGB frames (\textbf{Appearance}) rigidly conditioned on this verified structural layout. By establishing a physically grounded, pixel-aligned skeleton before rendering, we effectively eliminate entanglement artifacts and resolve the multi-agent control dilemma.

Furthermore, to elevate the model from passive instruction-following to active physical reasoning, we introduce a \textbf{Masked Point Recovery} training strategy. By randomly occluding or truncating input agent trajectories and compelling the network to reconstruct the complete underlying dynamic depth sequence, the model learns to infer causal continuations consistent with scene dynamics within its training distribution.

While designed primarily for autonomous driving (evaluated on Waymo, Driving Dojo, and YouTube), the structural integrity of our paradigm exhibits strong cross-domain robustness.
In summary, our main contributions to structured video generation are fourfold:

\noindent\textbullet~ \textbf{Bridging the Generative Gap.} We identify and resolve the domain gap between simple human input and dense physical output. Our ``Point-Shape-Appearance'' hierarchy enables precise, low-barrier multi-agent control without requiring dense 3D bounding boxes.

\noindent\textbullet~ \textbf{Decoupling Ego and Object Motion.} By introducing dynamic depth as the intermediate ``Shape'' stage, we structurally decouple camera motion from dynamic objects in 3D space, eliminating the entanglement artifacts common in previous sparse-control methods.

\noindent\textbullet~ \textbf{Active Physical Reasoning.} We introduce Masked Point Recovery, which forces the model to reconstruct dynamic shapes from interrupted trajectories, learning to infer causal continuations from partial observations within its training distribution.

\noindent\textbullet~ \textbf{Robust Generalizability.} We demonstrate that our framework maintains stable, physically coherent generation not only in driving scenarios, but also in highly complex rigid-body collisions (Physion) and embodied robotic manipulation (Jaco Play).
\section{Related Work}
\label{sec:related}

\textbf{Video Diffusion Models.} Early video generation primarily extended UNet-based latent diffusion models (LDMs)~\cite{rombach2022high} with temporal attention mechanisms~\cite{guo2023animatediff, wang2023modelscope, cerspense2023zeroscope, chen2023videocrafter1, chen2024videocrafter2, zhang2024moonshot, zhang2023show}. Recently, the field shifted toward large-scale Diffusion Transformers (DiT) and Flow Matching~\cite{lipman2022flow}, with foundational models~\cite{sora2024, yang2024cogvideox, kong2024hunyuanvideo, wan2025wan21} significantly improving spatiotemporal coherence.

\textbf{Controllable Motion and World Models.} Bridging intent and generation, prior works enable control via trajectories~\cite{yin2023dragnuwa, wu2024draganything}, keypoints, or flow~\cite{cheng2023sgi2v, niu2024mofa}. In parallel, autonomous driving world models~\cite{wang2023drivedreamer, zhao2024drivedreamer2, li2023drivingdiffusion, hu2023gaia, wang2024worlddreamer, yang2024genad} emerged to simulate driving. Recent advances integrate 3D geometry to enhance consistency~\cite{ren2025gen3c, yang2026neoverse, lu20254d, chen2025geodrive}. Concurrent work STANCE~\cite{chen2025stance} also adopts sparse-to-dense control, encoding instance cues as attention tokens for joint RGB generation. In contrast, Motion Forcing expands sparse points into an explicit depth intermediate under independent timesteps, providing a geometric scaffold that hard-constrains the RGB stage rather than soft-conditioning it.

\textbf{Physically Coherent Video Generation.} Standard models often violate physical laws~\cite{motamed2025generative, bansal2024videophy}. Prior solutions integrate physics engines~\cite{liu2024physgen}, leverage LLMs for reasoning~\cite{zhang2025think, zhang2025videorepa}, or embed latent priors~\cite{kundur2026physvideo}. However, entangling dynamics and appearance remains a bottleneck, as textures often override structure. We resolve this by explicitly decoupling physical reasoning from rendering.
\section{Method}
\label{sec:methods}

Generating controllable structured videos (e.g., autonomous driving) requires jointly reasoning about two intertwined aspects: \textit{where things move} (physical dynamics) and \textit{what things look like} (visual rendering). Mapping sparse controls $\mathbf{P}$ directly to dense pixels $\mathbf{V}$ conflates these challenges, forcing models to simultaneously resolve 3D geometry and local textures. We address this by introducing \textbf{Depth} as an intermediate representation, decomposing generation into \textit{Physical Reasoning} ($\mathbf{P}, \mathbf{C} \to \mathbf{D}$) and \textit{Neural Rendering} ($\mathbf{D} \to \mathbf{V}$), where $\mathbf{C}$ denotes camera motion. Crucially, sharing a single diffusion backbone enables geometric feature reuse and avoids representation misalignment.

We formalize the representations (\cref{sec:methods-pre}), describe depth warping for camera motion (\cref{sec:methods-camera}), the hierarchical forcing formulation (\cref{sec:methods-forcing}), the masked point recovery strategy (\cref{sec:methods-mask}), and inference (\cref{sec:methods-infer}).

\begin{figure}[!t]
    \centering
    \includegraphics[width=\linewidth]{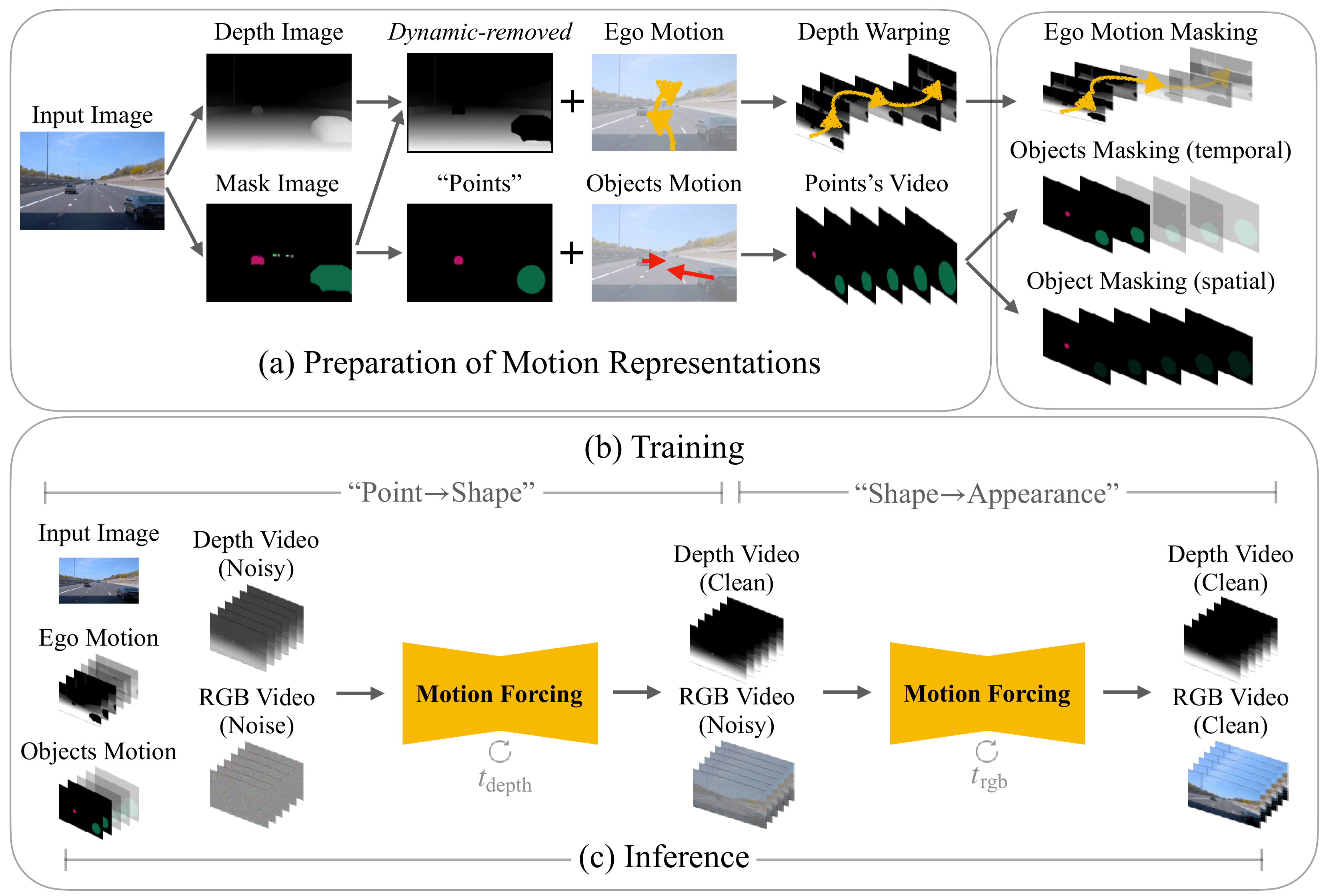}
    \caption{\textbf{Overview of the Motion Forcing framework.} \textbf{(a) Preparation of Motion Representations:} Input control signals are processed and subjected to spatial and temporal masking before being fed into the model. \textbf{(b) Training:} The model is trained by randomly sampling between two independent stages: "Point $\to$ Shape" (depth generation) and "Shape $\to$ Appearance" (RGB rendering). \textbf{(c) Inference:} The generation follows a complete two-stage hierarchical process, sequentially mapping sparse points to structural depth (Shape), and finally to the target RGB video (Appearance).}
    \label{fig:overview}
\end{figure}

\subsection{Preliminaries}
\label{sec:methods-pre}

Given a reference image $\mathbf{I}_0$ and a sequence of sparse control signals, our goal is to generate a physically plausible driving video $\mathbf{V} = \{\mathbf{I}_t\}_{t=0}^{T}$ that faithfully follows the prescribed vehicle motions and camera trajectory. We formulate this as a hierarchical mapping across three levels of scene description:

\textbf{Point} $\mathbf{P} = \{\mathbf{P}_t\}_{t=0}^{T}$: Sparse control signals for dynamic agents, abstracted as maximum inscribed circles (centroid and radius). The centroid governs planar motion, while the radius implicitly encodes depth via perspective projection, offering procedural flexibility. \textbf{Shape (Depth)} $\mathbf{D} = \{\mathbf{D}_t\}_{t=0}^{T}$: Intermediate structural representation of dense depth maps. Depth captures continuous 3D scene geometry---surface distances, spatial relationships, and occlusion ordering---bridging control-level reasoning and pixel-level rendering. \textbf{Appearance} $\mathbf{V} = \{\mathbf{I}_t\}_{t=0}^{T}$: The target RGB video containing high-frequency visual details (texture, lighting, materials).

In addition to agent-level control, driving scenarios require explicit camera motion specification. We denote the camera trajectory as $\mathbf{C} = \{(\mathbf{R}_t, \mathbf{t}_t, \mathbf{K}_t)\}_{t=0}^{T}$, comprising per-frame extrinsics $[\mathbf{R}_t | \mathbf{t}_t] \in \mathbb{R}^{3 \times 4}$ and intrinsics $\mathbf{K}_t \in \mathbb{R}^{3 \times 3}$. During training, these are estimated by VGGT~\cite{wang2025vggt}. We next describe how $\mathbf{C}$ is encoded into a form compatible with our depth-centric framework.

\subsection{Camera Motion via Depth Warping}
\label{sec:methods-camera}

A standard approach to injecting camera motion into a diffusion model is to embed the extrinsic parameters $(\mathbf{R}_t, \mathbf{t}_t)$ and intrinsics $\mathbf{K}_t$ as global conditioning vectors via cross-attention or adaptive normalization. However, this parametric strategy faces two critical challenges in the driving domain.

First, \textbf{entanglement under limited data diversity.} Injecting camera pose as a low-dimensional embedding alongside other conditions tends to entangle camera motion with scene content. This is exacerbated in driving datasets, where camera trajectories correlate strongly with scene layout and agent behavior, making factorization exceptionally difficult without massive data.

Second, \textbf{insufficient spatial precision.} Autonomous driving demands geometrically precise 3D camera manipulation. A parametric embedding compresses the 6-DoF transformation into a vector that the network must internally re-expand into dense displacements---a lossy bottleneck poorly suited for pixel-aligned processing.

To simultaneously address these issues, we propose representing camera motion as \textbf{warped depth maps}. By geometrically warping the first-frame depth $\mathbf{D}_0$ according to the target camera pose, we produce a dense, pixel-aligned conditioning signal. This representation is (i) \textit{structurally decoupled} from appearance and layout conditions by construction, as it encodes only geometric camera effects, and (ii) spatially precise, making the full 6-DoF transformation explicit at every pixel location. Furthermore, since our intermediate representation is also depth-based, encoding camera motion in the same modality establishes a shared geometric vocabulary. This is significantly easier for the network to internalize than bridging the gap between abstract pose parameters and pixel-space effects.

Formally, given the first-frame depth $\mathbf{D}_0$ and camera poses $\{(\mathbf{R}_t, \mathbf{t}_t, \mathbf{K}_t)\}_{t=0}^{T}$, we construct the camera motion conditioning $\mathbf{W} = \{\mathbf{W}_t\}_{t=0}^{T}$ by unprojecting the first-frame depth into 3D space, transforming it to each target camera's perspective, and forward-splatting the depth values onto the target image plane. Because this geometric warping assumes a static scene, dynamic agents in $\mathbf{D}_0$ may leave localized ``ghosts'' at their initial positions in $\mathbf{W}_t$. However, we provide the concurrent sparse point trajectories $\mathbf{P}$ as an overriding control signal. The attention mechanism of the DiT effectively learns to utilize $\mathbf{W}_t$ exclusively as a rigid background constraint while attending to $\mathbf{P}$ to guide dynamic agent rendering, successfully resolving this potential spatial conflict. The detailed mathematical formulation for this projection and splatting process is provided in Appendix~\ref{app:depth-warp}.

\subsection{Motion Forcing: Unified Hierarchical Diffusion}
\label{sec:methods-forcing}

Rather than cascading separate models for reasoning and rendering, we propose a \textbf{single unified diffusion model} $\epsilon_\theta$ that handles both stages within a shared 3D DiT backbone~\cite{yang2024cogvideox}. The key mechanism enabling this is \textbf{dual independent diffusion timesteps}: $\tau_d$ controls the noise level of the depth latents, and $\tau_v$ controls the noise level of the video latents.

Let $\mathbf{z}_{\tau_d}^{d}$ and $\mathbf{z}_{\tau_v}^{v}$ denote the VAE-encoded latents of $\mathbf{D}$ and $\mathbf{V}$, perturbed by Gaussian noise at levels $\tau_d$ and $\tau_v$ respectively. These latent streams are concatenated temporally and processed jointly by the backbone. While doubling the sequence length $T$ increases the self-attention computational cost during training, this overhead is isolated to the joint training phase. At inference time, because the generation is strictly decoupled into two distinct passes (first predicting depth, then rendering RGB), the model reverts to the standard $O(T^2)$ sequence length per stage, ensuring highly scalable generation. Each stream is modulated by its specific timestep embedding via adaptive layer normalization (AdaLN), enabling disentangled denoising control. The unified objective is:
\begin{equation}
\label{eq:loss}
\mathcal{L} = \mathbb{E}_{\mathbf{D}, \mathbf{V}, \tau_d, \tau_v, \epsilon} \left[ \left\| \epsilon - \epsilon_\theta\!\left(\mathbf{z}_{\tau_d}^{d} \oplus \mathbf{z}_{\tau_v}^{v},\; \tau_d,\; \tau_v,\; \mathbf{I}_0,\; \mathbf{P},\; \mathbf{W}\right) \right\|^2 \right],
\end{equation}
where $\oplus$ denotes temporal concatenation. The conditioning inputs---reference image $\mathbf{I}_0$, point control $\mathbf{P}$, and camera depth $\mathbf{W}$---are each VAE-encoded and concatenated along the channel dimension of the noisy input.

\subsubsection{Dual Adaptive Layer Normalization}
To enable a single set of transformer blocks to simultaneously serve two denoising tasks, we extend the standard AdaLN mechanism. For each block, the depth and video halves of the hidden state sequence receive \textit{different} scale and shift parameters derived from their respective timestep embeddings $\tau_d$ and $\tau_v$:
\begin{equation}
\label{eq:adaln}
\mathbf{h}'_{1:T} = \text{LN}(\mathbf{h}_{1:T}) \odot \left[\gamma(\tau_d) \| \gamma(\tau_v)\right] + \left[\beta(\tau_d) \| \beta(\tau_v)\right],
\end{equation}
where $\|$ denotes concatenation along the temporal (sequence) axis, and $\gamma, \beta$ are learned affine functions. At the output layer, norm\_out is applied sequentially with both embeddings to produce the final prediction. This design allows the backbone attention layers to share representations while the normalization provides task-specific modulation.

\subsubsection{Forcing Strategy}
Inspired by Diffusion Forcing~\cite{chen2024diffusion}, we employ a \textbf{stochastic mode-switching} strategy. Each training iteration forces the model to solve exactly one sub-problem by fixing the non-target timestep:

\textbf{Mode~I: Physical Reasoning} (Point $+$ Camera $\to$ Shape). Set $\tau_v = T_{\max}$ (pure noise) and sample $\tau_d \sim \mathcal{U}\{0, \ldots, T_{\max}\!-\!1\}$. Deprived of RGB cues, the model reconstructs $\mathbf{D}$ from the sparse control $\mathbf{P}$, camera depth $\mathbf{W}$, and reference $\mathbf{I}_0$. This mode cultivates physical reasoning: inferring 3D structure evolution, resolving occlusions, and predicting dynamic depth changes from minimal control inputs. The loss is $\mathcal{L}_{\text{reason}} = \mathbb{E}[\| \epsilon_d - \hat{\epsilon}_d \|^2]$ s.t. $\tau_v = T_{\max}$.

\textbf{Mode~II: Neural Rendering} (Shape $\to$ Appearance). Set $\tau_d = 0$ (ground truth) and sample $\tau_v \sim \mathcal{U}\{0, \ldots, T_{\max}\!-\!1\}$. Conditioned on perfect geometry, the model focuses on hallucinating consistent textures, lighting, and material properties. The loss is $\mathcal{L}_{\text{render}} = \mathbb{E}[\| \epsilon_v - \hat{\epsilon}_v \|^2]$ s.t. $\tau_d = 0$.

By alternating between modes, $\epsilon_\theta$ learns to function as both a physics engine and a renderer within a shared latent space.

\subsection{Masked Point Recovery for Physical Reasoning}
\label{sec:methods-mask}

To explicitly enforce causal reasoning and active physical understanding, we introduce \textbf{Masked Point Recovery}. A critical secondary objective of this strategy is to bridge the inherent gap between dense training trajectories and the sparse, partial inputs typical of real-world inference. By systematically corrupting input motion controls while strictly supervising on the complete intermediate depth video $\mathbf{D}$, we compel the model to internalize physical laws (e.g., inertia, object permanence) and hallucinate missing geometries. As shown in \cref{fig:overview}, we employ three distinct masking strategies:

\textbf{Temporal Ego Masking:} We randomly truncate ego-motion conditioning. Camera signals after frame $t = \lfloor \tau_{\text{ego}} T \rfloor$, with a cutoff ratio $\tau_{\text{ego}} \sim \mathcal{U}(0.3, 1.0)$, are masked. The model must extrapolate the subsequent ego-trajectory from initial momentum and scene layout. \textbf{Temporal Object Masking:} Similarly, we apply a temporal mask to object control points $\mathbf{P}$ beyond a sampled cutoff $\tau_{\text{obj}} \sim \mathcal{U}(0.3, 1.0)$, compelling the model to predict ongoing object motion from initial velocities. \textbf{Spatial Object Masking:} To model object permanence and implicit interactions, we independently drop the entire trajectory of each object $i$ with probability $p_{\text{drop}}$. The model must deduce the missing agent's physical presence and trajectory from the target depth $\mathbf{D}$ and the reactive behaviors of visible agents.

\subsection{Inference: Two-Stage Hierarchical Generation}
\label{sec:methods-infer}

In inference, the dual-timestep formulation enables sequential chaining of the learned capabilities. Note that the target camera poses $\mathbf{C}$ required for depth warping can be flexibly extracted from reference videos or procedurally generated, as detailed in Appendix~\ref{app:inference-camera}:

\textbf{Stage~1: Depth Generation (Physical Reasoning).} Initialize $\mathbf{z}^d, \mathbf{z}^v \sim \mathcal{N}(\mathbf{0}, \mathbf{I})$. Execute the DDIM sampling loop targeting only the depth stream ($\tau_v$ fixed at $T_{\max}$). The model operates in Mode~I, synthesizing a clean depth latent $\hat{\mathbf{z}}^d$ from the control inputs $\mathbf{P}$ and camera motion $\mathbf{W}$. Throughout this stage, the RGB noise $\mathbf{z}^v$ is held constant.

\textbf{Stage~2: Appearance Synthesis (Neural Rendering).} Fix the depth latent to $\hat{\mathbf{z}}^d$ (setting $\tau_d = 0$) and re-initialize $\mathbf{z}^v$ with fresh noise. Run a second DDIM loop targeting the video stream, with the model operating in Mode~II. The generated depth serves as a fixed geometric blueprint for rendering.

This hierarchical pipeline offers \textbf{interpretability}: the intermediate depth $\hat{\mathbf{D}}$ provides a verifiable 3D scene layout that users can inspect before committing to the expensive rendering stage. It also provides a natural editing interface---users may modify the predicted depth (e.g., removing or repositioning agents) and re-render with consistent appearance.
\section{Experiments}
\label{sec:exp}

\begin{figure}[!t]
    \centering
    \includegraphics[width=\linewidth]{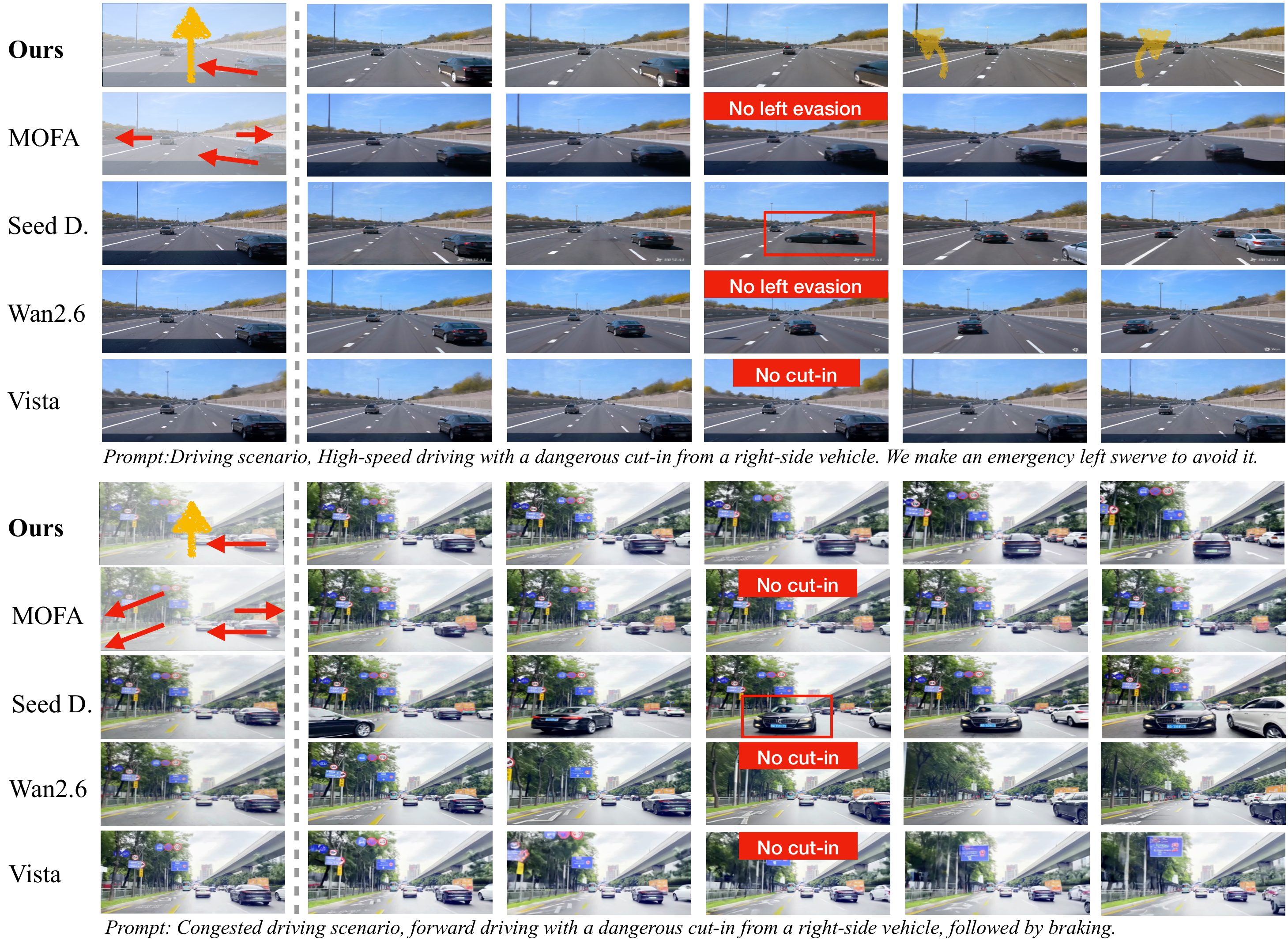}
    \caption{\textbf{Comparison with state-of-the-art models in synthesizing cut-in and evasive driving scenarios.} While the leading text-conditioned models (Seed Dance2.0, Wan2.6, Vista) rely on text prompts, trajectory-controllable baselines like MOFA-Video require object trajectories coupled with scene trajectories to simulate camera movement. Our method \textbf{Motion Forcing}, however, effectively captures these complex dynamics given only the initial ego and object motion trajectories.}
    \label{fig:comparison_drive}
    \vspace{0.5em}
    \centering
    \captionof{table}{Quantitative results on the Waymo dataset. Although our geometry-first approach yields a higher FVD compared to massively pretrained closed-source models, this is a deliberate tradeoff that significantly improves motion consistency (FVMD) and physical plausibility (Physics-IQ). One-stage$^*$ refers to the single-stage version of Motion Forcing, which directly generates RGB frames without intermediate depth.}
    \vspace{1em}
    \label{tab:waymo}
    \scalebox{1}{
    \begin{tabular}{lcccccc}
      \toprule
      \textbf{Methods} & MOFA-Video & Seed Dance2.0 & Wan2.6 & Vista & One-stage$^*$  & \textbf{Ours} \\
      \midrule
      \textbf{FVD$\downarrow$}  & 272.6 & \textbf{112.5} & 118.3 & 147.2 & 152.4 & 157.8 \\
      \textbf{FVMD$\downarrow$} & 421.3 & 345.6 & 316.2 & 323.1 & 218.4 & \textbf{205.2} \\
      \textbf{Physics-IQ$\uparrow$} & 21.6 & 30.5 & 31.2 & 28.4 & 30.7 & \textbf{33.2} \\
      \bottomrule
    \end{tabular}}
\end{figure}

\subsection{Experimental Setup}
\textbf{Training Details.} We fine-tune CogVideoX1.5-5B-I2V~\cite{yang2024cogvideox}. Comprehensive training details and hyperparameters are provided in Appendix~\ref{app:training-details}.
\textbf{Datasets.} We primarily evaluate on \textbf{autonomous driving} using sequences from Waymo~\cite{sun2020scalability}, Driving Dojo~\cite{wang2024drivingdojo}, and YouTube. To validate generality, we additionally evaluate on \textbf{Physion}~\cite{bear2021physion} (rigid-body physics) and \textbf{Jaco Play}~\cite{dass2023jacoplay} (robotic manipulation).
\textbf{Evaluation Metrics.} We report results on 100 Waymo~\cite{sun2020scalability} test videos using: (1) FVD~\cite{unterthiner2019accurategenerativemodelsvideo} for distributional similarity, (2) FVMD~\cite{liu2024fr} for temporal motion coherence, and (3) Physics-IQ~\cite{motamed2025generative} for physical plausibility.

\subsection{Comparisons with the State-of-the-art Methods}
\label{sec:exp-comparison}

\begin{figure}[!t]
    \centering
    \includegraphics[width=\linewidth]{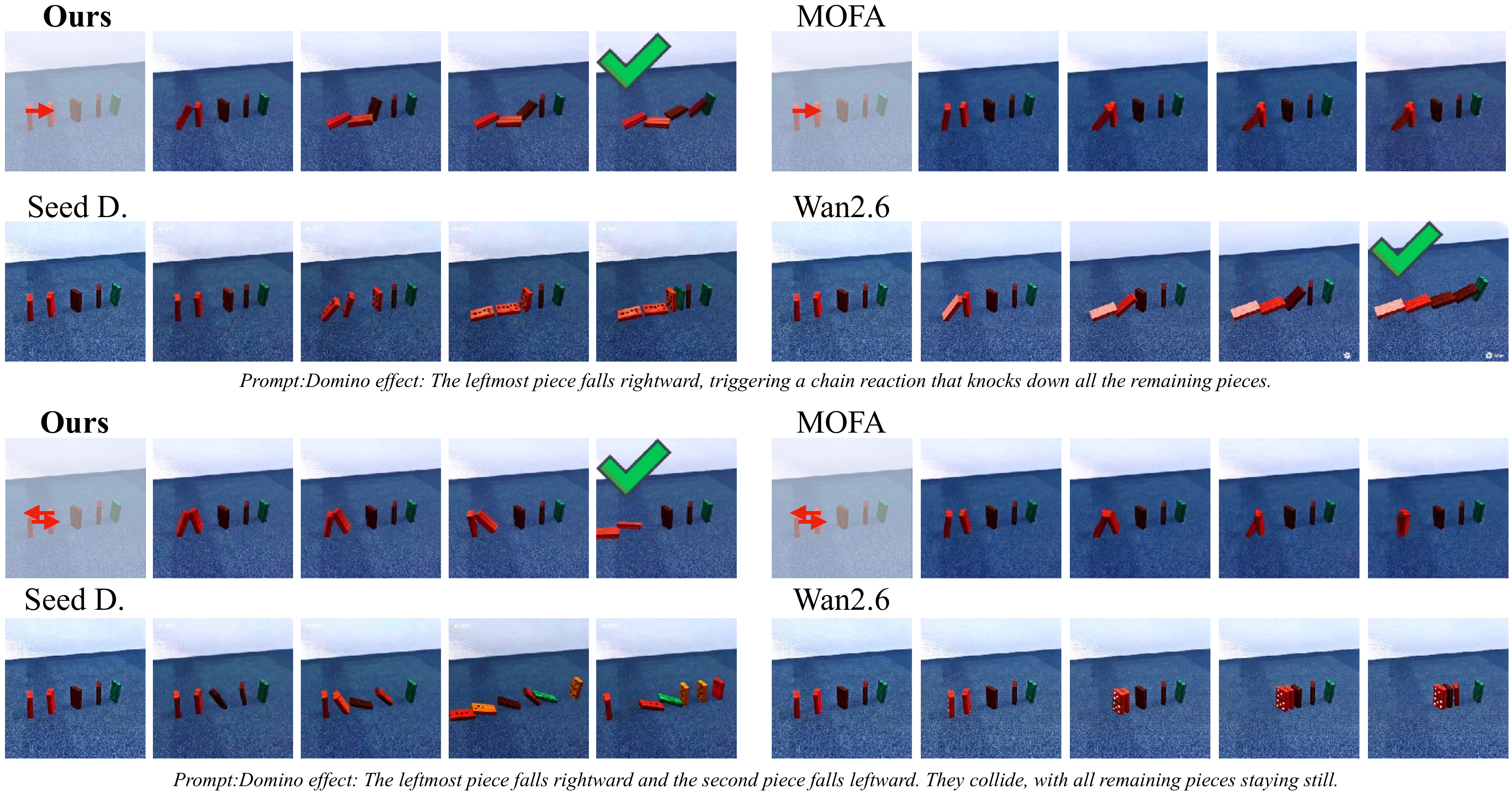}
    \caption{\textbf{Validation of physical capabilities.} \textcolor{green}{\checkmark} indicate physically coherent. MOFA-Video is fine-tuned on our exact dataset for fairness. As the control commands become more complex, only our method maintains stable and physically coherent generation.}
    \label{fig:comparison_domi}
\end{figure}
\begin{figure}[htb]
    \centering
    \includegraphics[width=\linewidth]{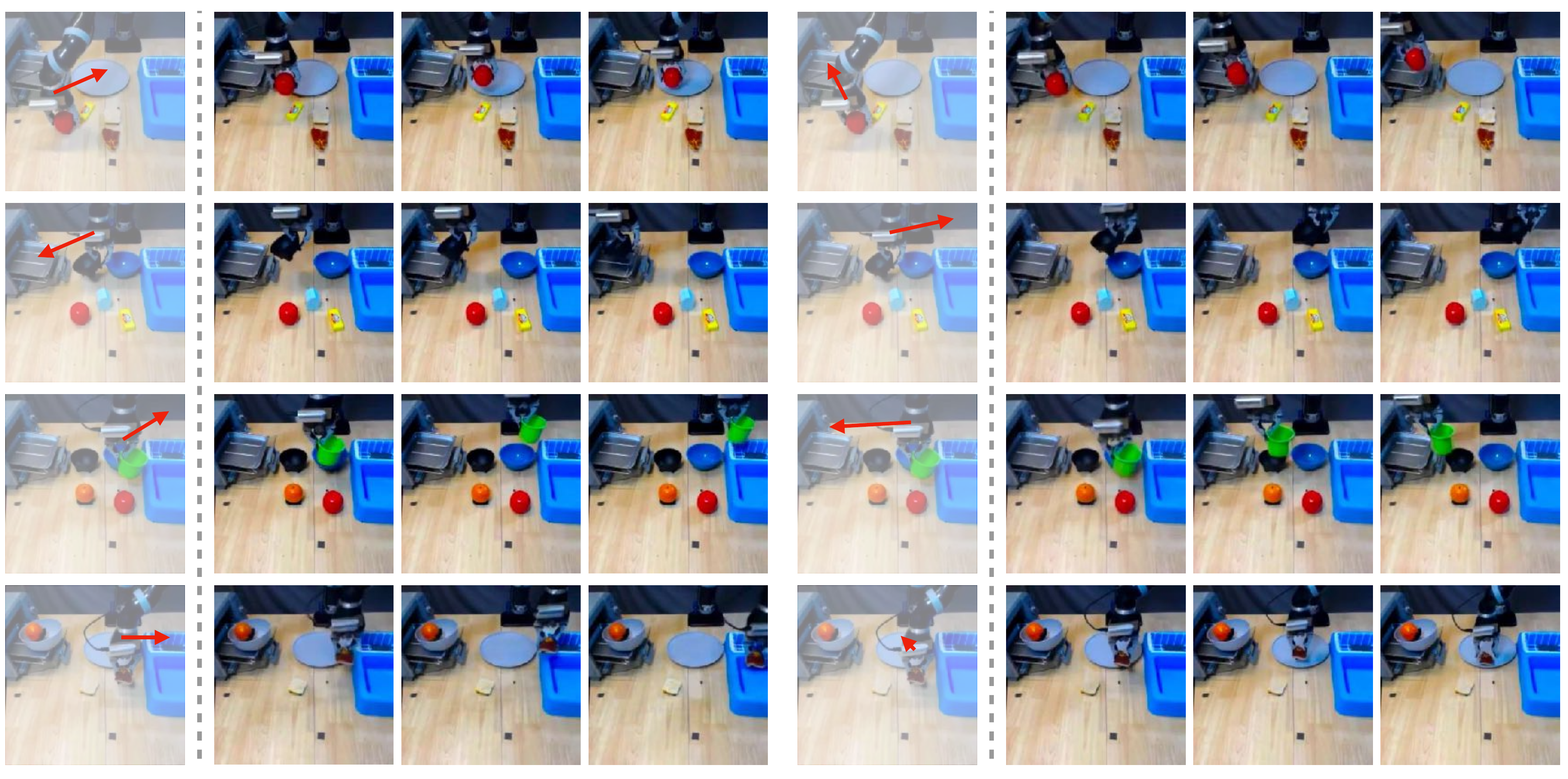}
    \caption{\textbf{Controllable robotic hand manipulation.} By providing different motion control inputs, our model can control the movement of the robotic arm and the grasped object in specified directions.}
    \label{fig:embodied}
\end{figure}

\noindent
\textbf{Real-world Driving Video Generation.}
As shown in \cref{tab:waymo}, we evaluate against MOFA-Video~\cite{niu2024mofa}, Vista~\cite{gao2024vista}, a one-stage variant, and text-conditioned I2V models (Seed Dance2.0~\cite{pang2024seed}, Wan2.6~\cite{wan2025wan21}). Map-conditioned generators MagicDrive~\cite{gao2023magicdrive} and DriveDreamer~\cite{wang2023drivedreamer} are excluded as they require dense HD maps and 3D bounding boxes, incompatible with our sparse-control premise. Text-conditioned baselines achieve lower FVD via massive pretraining scale, yet fall significantly behind on FVMD and Physics-IQ, confirming text prompts cannot substitute for explicit trajectory control. The one-stage variant likewise yields inferior physical metrics, confirming intermediate depth is crucial. Qualitatively (\cref{fig:comparison_drive}), text-conditioned models struggle with precise dynamics, and MOFA-Video degrades during multi-agent interactions due to coarse flow splatting, whereas our method faithfully synthesizes lane changes and evasive maneuvers. An AD-specific evaluation on nuScenes clips (\cref{tab:nuscenes-ad-proxy}) further confirms our geometry-first approach achieves significantly lower trajectory error and collision rates.

\begin{figure}[!t]
  \centering
  \includegraphics[width=\linewidth]{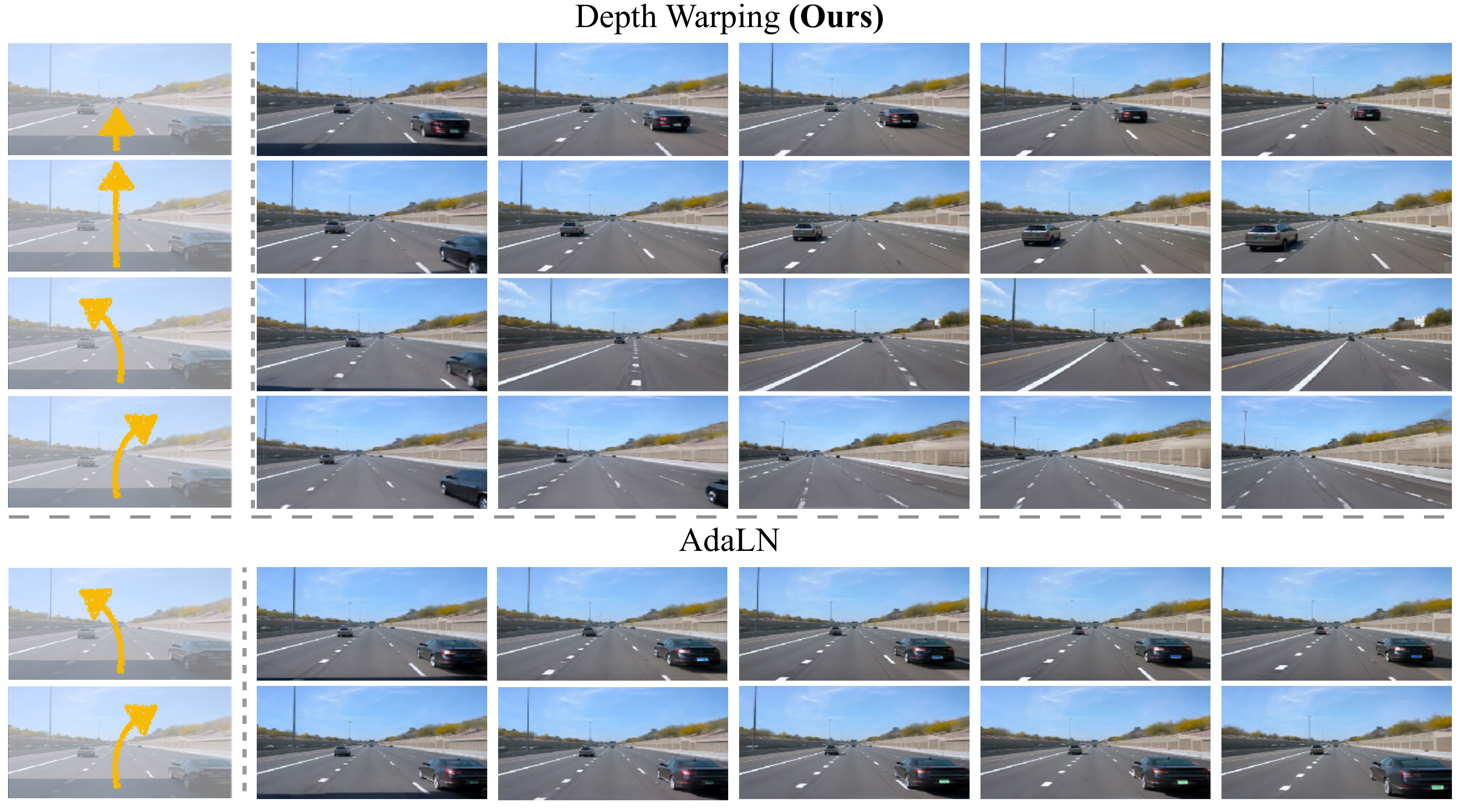}
    \caption{\textbf{Qualitative comparison of ego motion control.} Compared to the baseline AdaLN approach, our \textbf{Depth Warping-based} method demonstrates significantly superior accuracy and flexibility in controlling both the direction and speed.}
  \label{fig:ego_motion}
\end{figure}

\noindent\textbf{Generalization to Physics Scenarios.}
To validate the generality of our framework beyond driving, we evaluate on the Physion~\cite{bear2021physion} dataset. As shown in \cref{fig:comparison_domi}, when physical interactions grow more complex (\eg, multi-object collisions in domino scenarios), MOFA-Video---even fine-tuned on our dataset---fails to maintain physical coherence, whereas Motion Forcing consistently produces plausible collision dynamics, confirming that our hierarchy effectively transfers to general physics scenarios.

\begin{table}[thb]
\centering
\begin{minipage}[t]{0.55\linewidth}
\centering
\caption{\textbf{Ablation studies} on the Waymo test set. Each row replaces one component of Motion Forcing with the specified alternative.}
\label{tab:ablation}
\vspace{0.25em}
\scalebox{0.8}{
\begin{tabular}{lccc}
\toprule
\textbf{Methods} & \textbf{FVD$\downarrow$} & \textbf{FVMD$\downarrow$} & \textbf{Physics-IQ$\uparrow$} \\
\midrule
Motion Forcing & 157.8 & \textbf{205.2} & \textbf{33.2} \\
w/o Mask Point Recovery & \textbf{142.0} & 233.8 & 27.5 \\
\midrule
\multicolumn{4}{l}{\textbf{Motion Representation}} \\
Segmentation & 167.0 & 228.8 & 29.7 \\
Optical Flow & 173.0 & 224.3 & 28.4 \\
\midrule
\multicolumn{4}{l}{\textbf{Object Motion Embedding}} \\
Softmax Splatting & 160.3 & 251.7 & 28.5 \\
\midrule
\multicolumn{4}{l}{\textbf{Ego Motion Encoding}} \\
AdaLN & 159.6 & 243.8 & 29.1 \\
\bottomrule
\end{tabular}}
\end{minipage}
\hfill
\begin{minipage}[t]{0.42\linewidth}
\centering
\caption{\textbf{AD Proxy Evaluation} on a 50-video nuScenes subset. We compare our geometry-first model against the text-conditioned Vista baseline. Our approach yields lower trajectory tracking errors (L2) and agent collision rates.}
\label{tab:nuscenes-ad-proxy}
\vspace{0.25em}
\scalebox{0.86}{
\renewcommand{\arraystretch}{1.2}
\begin{tabular}{lcc}
\toprule
\textbf{Metric} & \textbf{Vista} & \textbf{Ours} \\
\midrule
\textbf{L2(1s)$\downarrow$} & 0.176 & \textbf{0.071} \\
\textbf{L2(2s)$\downarrow$} & 0.389 & \textbf{0.194} \\
\textbf{L2(3s)$\downarrow$} & 0.467 & \textbf{0.289} \\
\textbf{Collision$\downarrow$} & 0.581 & \textbf{0.498} \\
\bottomrule
\end{tabular}}
\end{minipage}
\end{table}

\noindent\textbf{Generalization to Robotic Manipulation.} We further validate on the Jaco Play~\cite{dass2023jacoplay} dataset by providing directional inputs to control robotic hand motions for grasping and object manipulation. As shown in \cref{fig:embodied}, our model flexibly guides the robotic arm and grasped objects in specified directions, confirming the flexibility of our point-based control primitive across domains.

\subsection{Ablation Study}
\label{sec:exp-ablation}
We conduct ablations on the Waymo test set to validate key design choices (\cref{tab:ablation}). The full two-stage model incurs a modest FVD increase over the one-stage variant (157.8 vs.\ 152.4): the depth-first decomposition geometrically constrains RGB generation, slightly trading distributional similarity for motion coherence and physical plausibility---a cost reflected in the substantially better FVMD and Physics-IQ. \textbf{Active Physical Reasoning:} Removing Masked Point Recovery modestly improves FVD by allowing passive overfitting to visual textures, but severely degrades FVMD and Physics-IQ, confirming its necessity for learning causal dynamics. \textbf{Intermediate Motion Representation:} Depth outperforms segmentation and optical flow by encoding continuous 3D geometry indispensable for motion reasoning. \textbf{Object Motion Embedding:} Replacing instance flow with softmax splatting~\cite{niu2024mofa} degrades FVMD and Physics-IQ, as coarse pixel-level warping struggles with multi-body dynamics. \textbf{Ego Motion Encoding:} Replacing depth warping with AdaLN markedly degrades motion coherence, confirming that spatially explicit, pixel-aligned 6-DoF conditioning is essential.

\section{Conclusion and Limitations}
\label{sec:conclu}
\label{sec:limitations}

We presented \textbf{Motion Forcing}, a unified diffusion backbone for controllable video generation via a Point$\,\to\,$Shape$\,\to\,$Appearance hierarchy using dual timesteps, depth-warped ego conditioning, and Masked Point Recovery. This depth-first decomposition separates physical reasoning from visual rendering, allowing the shared backbone to reuse geometric features. Across driving, physics, and robotics, it significantly improves motion coherence and physical plausibility under a sparse control.

\noindent\textbf{Broader Impacts.} While beneficial for driving simulation, robotics, and content creation, it could be misused to generate misleading synthetic footage; responsible deployment is advised.

\noindent\textbf{Limitations.} We do not surpass large closed-source models in raw FVD due to massive pretraining scale. Severe occlusion, dense traffic, and ID switches can disrupt sparse point control or depth ordering. \textbf{See supplementary for representative failure cases.}

\clearpage
\small
\bibliographystyle{plainnat}
\bibliography{main}

\normalsize
\clearpage
\appendix
\section{Additional Implementation Details}
\label{app:impl}

\subsection{Mathematical Formulation of Depth Warping}
\label{app:depth-warp}
As introduced in the main text, we construct the camera motion conditioning $\mathbf{W} = \{\mathbf{W}_t\}_{t=0}^{T}$ by unprojecting the first-frame depth into 3D, transforming it to each target camera, and forward-splatting:
\begin{equation}
\label{eq:warp}
\mathbf{W}_t = \text{Splat}\!\left(\mathbf{D}_0,\; \Pi_t \circ \Pi_0^{-1}(\mathbf{u}, \mathbf{D}_0(\mathbf{u}))\right).
\end{equation}
Concretely, for each pixel $\mathbf{u} = (u, v)$ in frame $0$, we first unproject to a world-space point:
\begin{equation}
\label{eq:unproject}
\mathbf{p}_{\text{world}} = \mathbf{R}_0^\top\!\left(\mathbf{D}_0(\mathbf{u}) \cdot \mathbf{K}_0^{-1}\begin{bmatrix} u \\ v \\ 1 \end{bmatrix} - \mathbf{t}_0\right),
\end{equation}
and then project it onto frame $t$'s image plane:
\begin{equation}
\label{eq:project}
\mathbf{u}_t = \pi\!\left(\mathbf{K}_t(\mathbf{R}_t \mathbf{p}_{\text{world}} + \mathbf{t}_t)\right),
\end{equation}
where $\pi$ denotes perspective division. Finally, the depth value $\mathbf{D}_0(\mathbf{u})$ is splatted to location $\mathbf{u}_t$ in the target frame.

\subsection{Camera Pose Construction During Inference}
\label{app:inference-camera}
During inference, the camera trajectory $\mathbf{C}$ required for Depth Warping can be constructed through multiple intuitive paradigms depending on the generation task. For video-to-video editing or style transfer scenarios where a reference video exists, we directly extract the camera poses from the source video using off-the-shelf pose estimators such as VGGT~\cite{wang2025vggt} or DA3~\cite{lin2025depth}. This provides a highly realistic, metric-accurate trajectory that mirrors the original video's ego-motion. For pure generation tasks from a single image, users can procedurally generate camera trajectories by defining a sequence of 6-DoF rigid body transformations. For example, a user can script a forward driving motion at a constant speed or simulate a lane change by mathematically computing the consecutive camera matrices $\mathbf{K}_t, \mathbf{R}_t, \mathbf{t}_t$. This flexibility ensures that Motion Forcing supports both data-driven tracking and explicitly scripted camera controls, maintaining the sparse-control paradigm without relying on dense scene representations.

\subsection{Training Details}
\label{app:training-details}
We fine-tune CogVideoX1.5-5B-I2V~\cite{yang2024cogvideox} on 8 NVIDIA A100 GPUs with DeepSpeed ZeRO Stage-2~\cite{rajbhandari2020zero} in \texttt{bfloat16} mixed precision. The total training process takes approximately one week. We use AdamW~\cite{loshchilov2019decoupled} with learning rate $1 \times 10^{-5}$, $(\beta_1, \beta_2) = (0.9, 0.95)$, weight decay $1 \times 10^{-4}$, and 100 warmup steps. Each sample consists of 33 frames at $320 \times 480$ resolution (stride 2). We train for 10 epochs with an effective batch size of 8.

\subsection{Point Construction and Control Interface}
\label{app:point-construction}
Motion Forcing uses the same point tensor format during training and inference.
For each dynamic object, we first obtain an instance mask or tracked object
region.  In datasets with annotated masks, such as Physion, we use the provided
mask directly; for driving videos, masks are obtained with an off-the-shelf
instance segmentation model and temporally associated with object tracks when
available.  Each mask is converted into the maximum inscribed circle, represented
by a center $(x_t^i,y_t^i)$ and radius $r_t^i$.  The centers specify image-plane
motion, while the radius provides a coarse perspective cue that helps distinguish
objects moving laterally from objects approaching or receding.

At inference time, the same representation can be produced from user sketches,
tracked objects, or scripted trajectories.  A user-provided arrow specifies the
desired displacement of an object center, while the radius can be kept fixed or
updated by a simple perspective rule.  This keeps the control interface sparse:
the model is not given a dense flow field or full segmentation video, and must
expand sparse point controls into dynamic depth before RGB rendering.

The main sources of point-construction error are imperfect masks, ID switches in
crowded scenes, and radius noise under heavy occlusion.  Such errors mostly
affect small non-motorized agents and densely overlapping objects.  Masked Point
Recovery partially exposes the model to missing or corrupted controls, but it is
not a substitute for accurate perception; the limitations section discusses
failure modes caused by severe tracking and occlusion ambiguity.

\subsection{Masked Point Recovery}
\label{app:mpr}
Masked Point Recovery is applied only during the Point-to-Shape training mode.
We randomly hide temporal suffixes of ego or object trajectories, or drop entire
object trajectories, while keeping supervision on the complete dynamic depth
sequence.  The objective asks the model to explain the hidden part of the scene
using visible motion, depth ordering, object persistence, and interaction cues.
This design is intended to encourage causal interaction inference in structured
domains, for example predicting how a nearby vehicle reacts to a cut-in or how a
rigid object moves after contact.  It should not be interpreted as learning
general physical laws independent of the training distribution.

\subsection{Shared Backbone}
\label{app:shared-backbone}
The unified backbone shares attention layers across the depth and RGB streams,
but the two streams are separated by independent diffusion timesteps and
task-specific AdaLN modulation.  During Point-to-Shape training, the RGB stream
is fixed at pure noise and the loss is applied to depth.  During
Shape-to-Appearance training, the depth stream is clean and the loss is applied
to RGB.  This mode switching prevents the model from solving both tasks through
a single undifferentiated denoising target, while still allowing the renderer to
reuse geometric features learned in the depth stage.

\section{Additional Qualitative Results}
\label{app:qualitative}

\subsection{Collected Driving Data}
\label{app:driving-data}
The YouTube driving subset focuses on highly interactive traffic scenes in which
ego motion or one vehicle's action influences surrounding vehicles or
pedestrians.  It contains more than 9,000 clips and roughly 200 hours of video
covering diverse weather, road layouts, urban scenes, and interaction patterns.
All case videos referenced in the main paper are provided in the supplementary
material.

\subsection{nuScenes AD Proxy Protocol and Depth Videos}
\label{app:nuscenes-ad-proxy}
We include an additional autonomous-driving proxy evaluation on 50 nuScenes
CAM\_FRONT clips.  The
reported trajectory metric first estimates camera poses from the generated RGB
video with DA3~\cite{lin2025depth}, then compares the recovered camera centers with the nuScenes
ego trajectory after alignment.  We report both Sim(3)-aligned and scale-only
L2, and use the Sim(3)-aligned L2 in the main table because monocular depth and
pose are only defined up to global similarity.  The collision number is a
lightweight any-object image-plane overlap proxy computed with instance
segmentation; it is intended as a diagnostic for visible object overlap rather
than a substitute for a closed-loop driving safety metric.

\subsection{Generation Pipeline and Point Control Robustness}
\label{app:point-controls}
\begin{figure}[p]
    \centering
    \includegraphics[width=\linewidth]{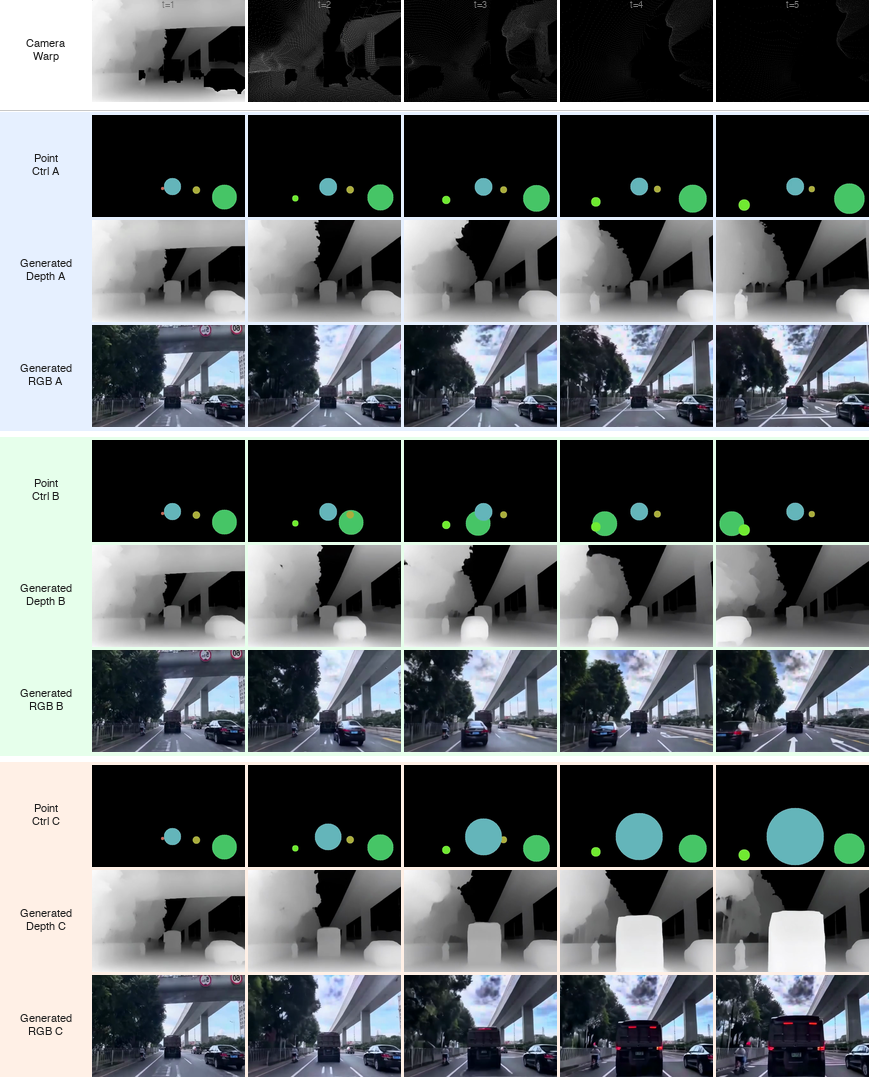}
    \caption{\textbf{Generation pipeline and point control robustness --- Scene~1.}
    Top row: shared camera-warped depth condition.  Rows 2--10: three point
    control variants (A/B/C), each showing point input, generated depth, and
    generated RGB (5 uniformly sampled frames per row).}
    \label{fig:appendix-point-controls-1}
\end{figure}

\begin{figure}[p]
    \centering
    \includegraphics[width=\linewidth]{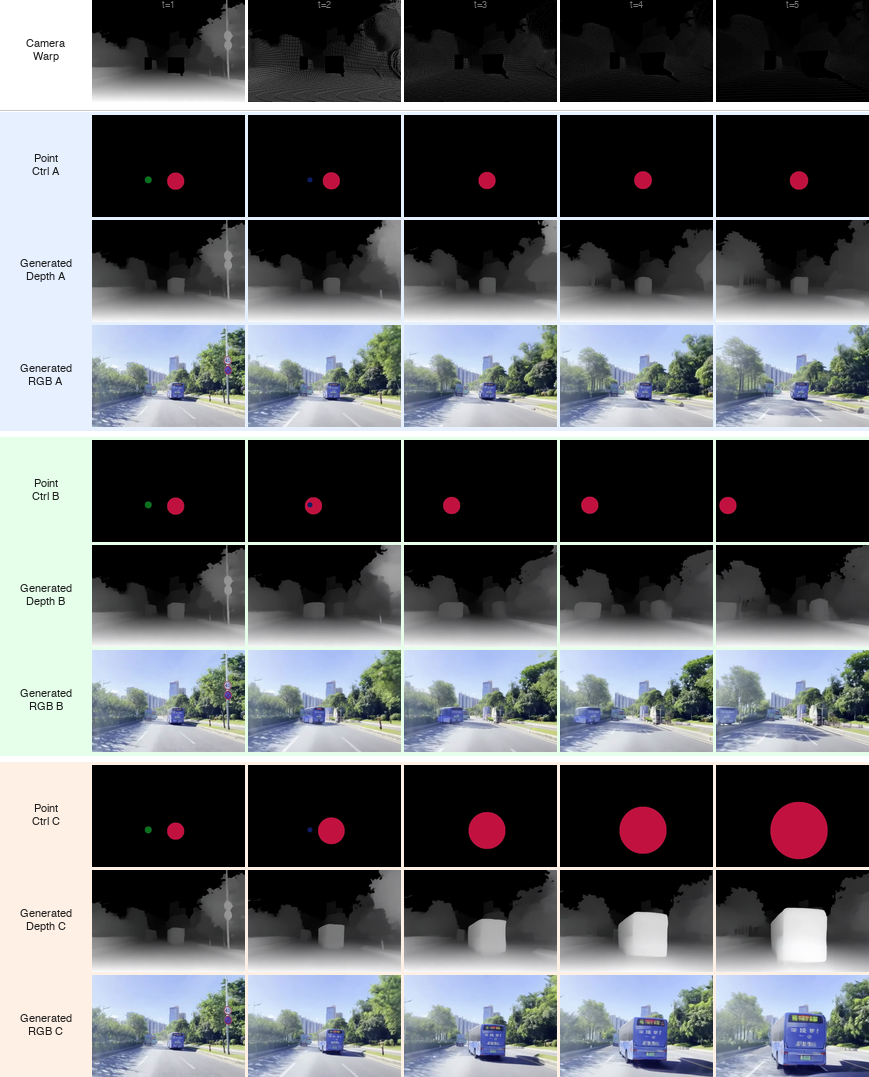}
    \caption{\textbf{Generation pipeline and point control robustness --- Scene~2.}
    Same layout as \cref{fig:appendix-point-controls-1}.  The model faithfully
    propagates structurally distinct trajectories through the depth stage before
    RGB rendering.}
    \label{fig:appendix-point-controls-2}
\end{figure}

\begin{figure}[p]
    \centering
    \includegraphics[width=\linewidth]{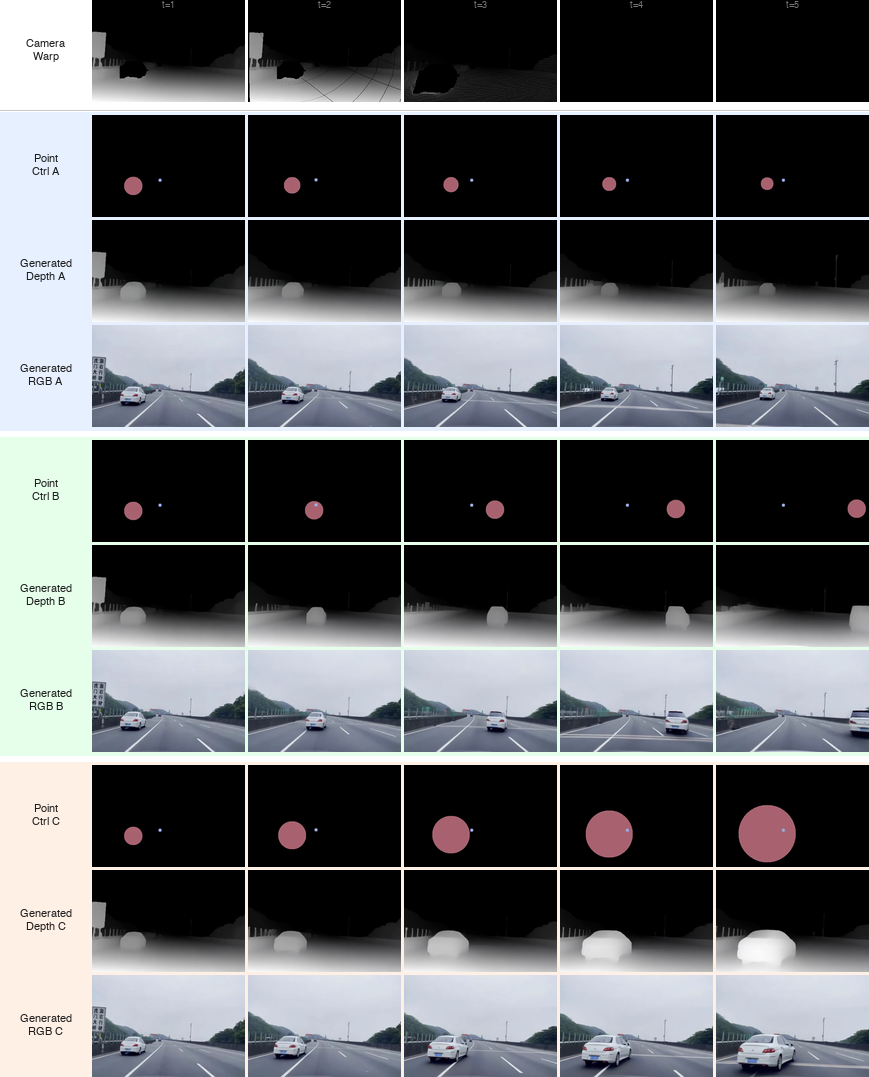}
    \caption{\textbf{Generation pipeline and point control robustness --- Scene~3.}
    Same layout as \cref{fig:appendix-point-controls-1}.  Across all three scenes
    the depth-first hierarchy cleanly separates physical reasoning from visual
    rendering, enabling robust control with diverse sparse inputs.}
    \label{fig:appendix-point-controls-3}
\end{figure}

Figures~\ref{fig:appendix-point-controls-1}--\ref{fig:appendix-point-controls-3}
visualize the complete two-stage generation pipeline across three driving scenes,
each with three distinct point control configurations (Ctrl A/B/C) applied to the
same camera motion.  The top row of each figure shows the shared camera-warped
depth condition $\mathbf{W}$.  Below, each group of three rows presents one
control variant: the sparse point input video $\mathbf{P}$, the generated
intermediate depth sequence $\mathbf{D}$, and the final rendered RGB video
$\mathbf{V}$.  The consistent depth structure and realistic appearance across all
three controls confirm that Motion Forcing reliably translates different sparse
trajectories into plausible scene dynamics without requiring retraining or
architectural changes.

\end{document}